\documentclass[11pt,letterpaper]{article}
\usepackage{emnlp2017}
\usepackage{times}
\usepackage{latexsym}

\emnlpfinalcopy

\def\emnlppaperid{***}

\usepackage{url}
\usepackage{color}
\usepackage{amsmath}
\usepackage{amssymb}
\usepackage{graphicx}
\usepackage{multirow}
\usepackage[linesnumbered,ruled,vlined]{algorithm2e}
\usepackage{algpseudocode}
\usepackage{todonotes}

\newcommand\QN[1]{\textrm{\color{red}#1}}
\newcommand\mbf[1]{\mathbf{#1}}
\newcommand\argmax[2]{\textrm{arg}\max_{#1}{#2}}
\newcommand{\event}[1]{\textit{\textbf{#1}}}
\newcommand{\rel}[1]{\textit{#1}}
\newcommand{\calx}{\mathcal{X}}
\newcommand{\caly}{\mathcal{Y}}
\newcommand{\real}{\mathbb{R}}
\newcommand{\ignore}[1]{}

\newcommand{\final}[1]{#1}

\usepackage{balance}
\title{ \vspace*{-0.5in}
{{\small \hfill EMNLP'17}\\
\vspace*{.25in}} A Structured Learning Approach to Temporal Relation Extraction}

\author{Qiang Ning$^1$ \and ~Zhili Feng $^2$ \and Dan Roth $^{1,2}$ \\
	$^1$Department of Electrical and Computer Engineering\\
	$^2$Department of Computer Science\\
	University of Illinois, Urbana, IL 61801\\
	{\tt \{qning2,zfeng6,danr\}@illinois.edu} \\}

\date{}

\begin{document}
	
	\maketitle
	
	\begin{abstract}
		Identifying temporal relations between events is an essential step towards natural language understanding. However, the temporal relation between two events in a story depends on, and is often dictated by, relations among other events. Consequently, effectively identifying  temporal relations between events is a challenging problem even for human annotators. 
		This paper suggests that it is important to take these dependencies into account while learning to identify these relations and proposes a structured learning approach to address this challenge. As a byproduct, this provides a new perspective on handling missing relations, a known issue that hurts existing methods. 
		As we show, the proposed approach results in significant improvements on the two commonly used data sets for this problem. 
		
		\ignore{Identifying temporal relations between events is an essential step towards natural language understanding. However, the temporal relation between two events in a story cannot be determined in isolation. The relations among many other events should be taken into account, making the effective identification of temporal relations a challenging problem even for human annotators. 
			This paper proposes a structured learning approach to address this challenge and a new perspective towards the vast majority of missing relations, the treatment of which is known to be a major issue that hurts existing methods. 
			The proposed approach results in significant improvements on two evaluation sets, compared to the state-of-the-art on this problem. 
		}
	\end{abstract}
\section{Introduction} \label{sec:intro}

\ignore{Can we describe the method under CCM?}

Understanding temporal information described in natural language text is a key component of natural language understanding \citep{mani2006machine, verhagen2007semeval, chambers2007classifying, bethard2007cu} \final{and, following} a series of TempEval (TE) workshops \citep{verhagen2007semeval,verhagen2010semeval,uzzaman2013TE3}, it has drawn \final{increased} attention.
Time-slot filling \citep{surdeanu2013overview,ji2014tackling}, storyline construction \citep{DoLuRo12,minard2015semeval}, clinical narratives processing \citep{JindalRo13d,bethard2016semeval}, and temporal question answering \citep{llorens2015semeval} are all explicit examples of temporal processing. 
	
The fundamental tasks in temporal processing, as identified in the TE workshops, are 1) time expression (the so-called ``timex'') extraction and normalization and 2) temporal relation (also known as TLINKs \citep{pustejovsky2003timeml}) extraction. 
%
While the first task has now been well handled by the state-of-the-art systems (HeidelTime \citep{strotgen2010heideltime}, SUTime \citep{chang2012sutime}, \final{IllinoisTime}~\citep{ZhaoDoRo12}, NavyTime \citep{chambers2013navytime}, UWTime \cite{lee2014context}, etc.) with end-to-end $F_1$ scores being around 80\%, the second task has long been a challenging one; even the top systems only achieved $F_1$ scores \final{of} around 35\% \final{in the TE workshops}.

The goal of the temporal relation task is to generate a \final{{\em directed temporal graph}} whose nodes represent temporal entities (i.e., events or timexes) and edges represent the TLINKs between them.
The task is challenging because \final{ it often requires global considerations -- considering the entire graph, the TLINK annotation is quadratic in the number of nodes and thus very expensive, and an overwhelming fraction of the temporal relations are missing in human annotation.}
\final{
In this paper, we propose a structured learning approach to temporal relation extraction, where local models are updated based on feedback from global inferences. The structured approach also gives rise to a semi-supervised method, making it possible to take advantage of the readily available unlabeled data. As a byproduct, this approach further provides a new, effective perspective on handling those missing relations.}

\begin{figure}[htbp!]
	\centering
	\includegraphics[width=0.45\textwidth]{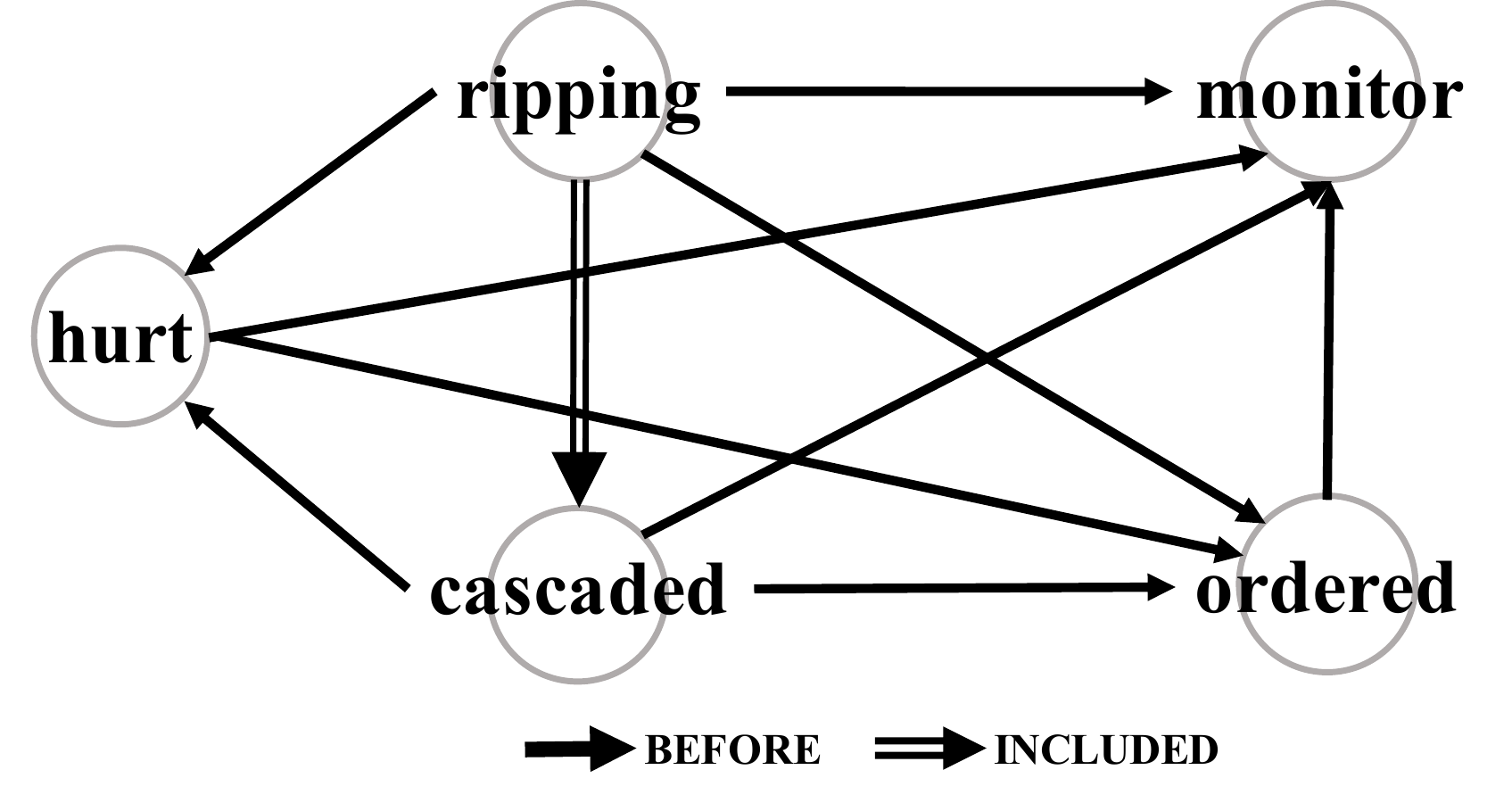}
	\caption{\small The desired event temporal graph for Ex1. Reverse TLINKs such as \event{hurt} is \rel{after} \event{ripping} are omitted for simplicity.}
	\label{fig:graph all}
\end{figure}

\final{In the common formulations, temporal relations are categorized into three types: the E-E TLINKs (those between a pair of events), the T-T TLINKs (those between a pair of timexes), and the E-T TLINKs (those between an event and a timex).}
\ignore{\final{The} temporal relation task requires \final{the identification of}
three types of temporal relations, \final{those} between a pair of events (E-E TLINKs), \final{between} a pair of timexes (T-T TLINKs), or \final{between an event} and \final{a} timex (E-T TLINKs).}
\final{While the proposed approach can be generally applied to all three types, this paper focuses on the majority type, i.e., the E-E TLINKs.}
For example, \final{consider the following snippet taken from the training set provided in the TE3 workshop}. We \final{want} to construct a temporal graph as in Fig.~\ref{fig:graph all} for the events in boldface in Ex1.
\begin{enumerate}
	\item [Ex1] \textit{\dots tons of earth \textbf{cascaded} down a hillside, \textbf{ripping} two houses from their foundations. No one was \textbf{hurt}, but firefighters \textbf{ordered} the evacuation of nearby homes and said they'll \textbf{monitor} the shifting ground.\dots}
\end{enumerate}

\begin{figure}[htbp!]
	\centering
	\includegraphics[width=0.45\textwidth]{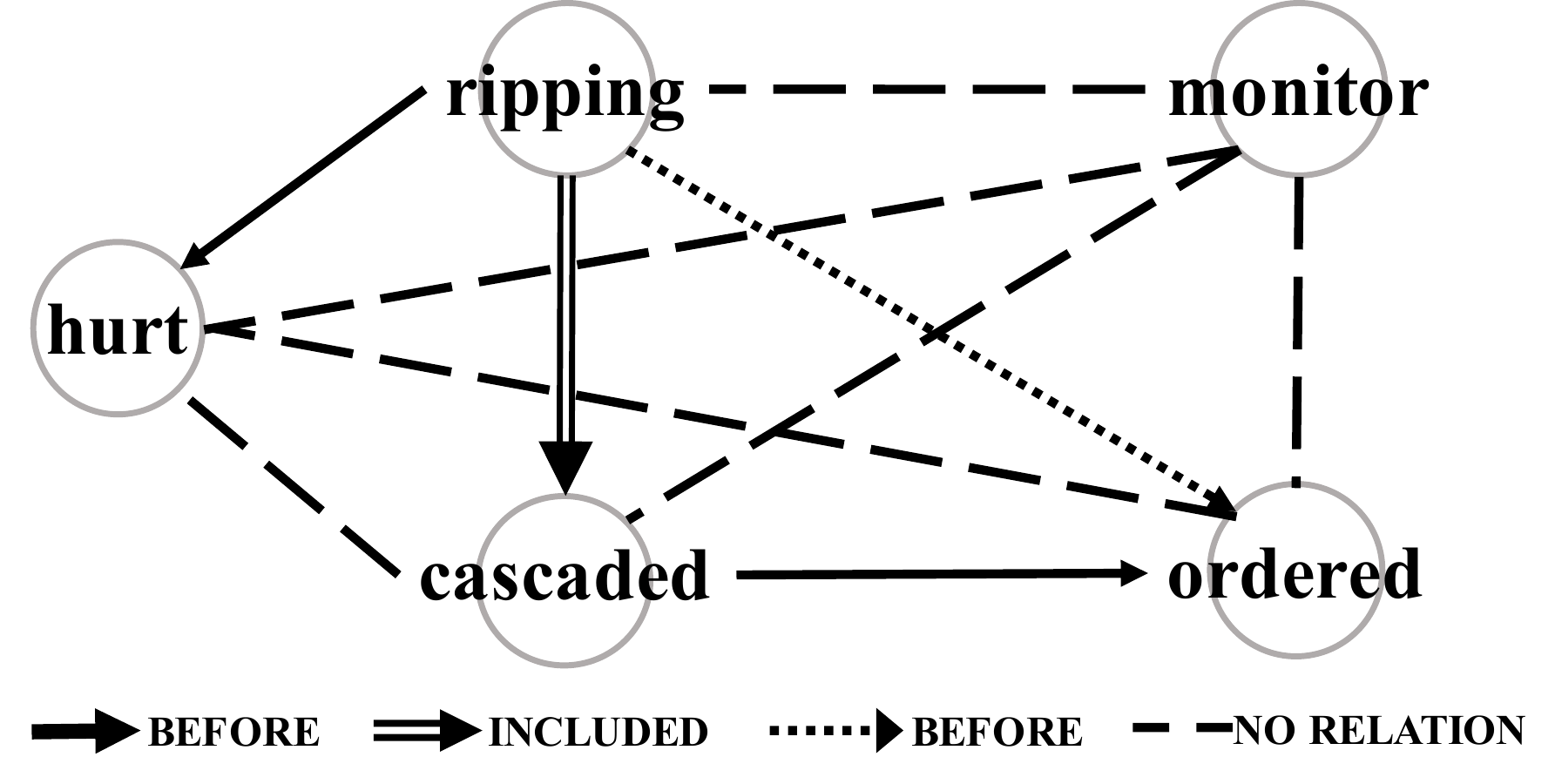}
	\caption{\small The human-annotation for Ex1 provided in TE3, where many TLINKs are missing due to the annotation difficulty. Solid lines: original human annotations. Dotted lines: TLINKs inferred from solid lines. Dashed lines: missing relations.}
	\label{fig:graph}
\end{figure}

As \final{discussed in} 
existing work \citep{verhagen2004times, BDLB06, mani2006machine, ChambersJu08}, the structure of a temporal graph is constrained by some rather simple rules:
\begin{enumerate}
	\item \textit{Symmetry}. For example, if \textit{A} is \textit{before B}, then \textit{B} must be \textit{after A}.
	\item \textit{Transitivity}. For example, if \textit{A} is \textit{before B} and \textit{B} is \textit{before C}, then \textit{A} must be \textit{before C}.
\end{enumerate}
This particular structure of a temporal graph (especially the transitivity structure) makes its nodes highly interrelated, as can be seen from Fig.~\ref{fig:graph all}. 
It is thus very challenging to identify the TLINKs between them, even for human annotators: The inter-annotator agreement on TLINKs is usually about 50\%-60\% \citep{mani2006machine}. Fig.~\ref{fig:graph} shows the actual human annotations provided by TE3. Among all the ten possible pairs of nodes, only three TLINKs \ignore{(i.e., \event{ripping} is \rel{before} \event{hurt}, \event{ripping} is \rel{included} in \event{cascaded}, and \event{cascaded} is \rel{before} \event{ordered})} were annotated.
Even if we only look at main events in consecutive sentences and at events in the same sentence, there are still quite a few missing TLINKs, e.g., the one between \event{hurt} and \event{cascaded} and the one between \event{monitor} and \event{ordered}.

\final{Early attempts 
by}~\citet{mani2006machine,chambers2007classifying, bethard2007timelines, verhagen2008temporal} studied \textit{local} methods -- learning models that make pairwise decisions between each pair of events. 
State-of-the-art local methods, including ClearTK \citep{bethard2013cleartk}, UTTime \citep{laokulrat2013uttime}, and NavyTime \citep{chambers2013navytime}, use better designed rules or more features such as syntactic tree paths and achieve better results.
However, \final{the decisions made by these (local) models are often globally inconsistent} (i.e., the symmetry and/or transitivity constraints are not satisfied for the entire temporal graph). 
Integer linear programming (ILP) methods~\cite{RothYi04} were used in this domain to enforce global consistency by several authors including \citet{BDLB06, ChambersJu08, DoLuRo12}, which formulated TLINK extraction as an ILP and showed that it improves over local methods for densely connected graphs. 
Since these methods perform inference (``I'') on top of pre-trained local classifiers (``L''), they are often referred to as {\textit{L+I}}~\citep{PRYZ05}.
In a state-of-the-art method, CAEVO \citep{chambers2014dense}, many hand-crafted rules and machine learned classifiers (called sieves therein) form a pipeline. The global consistency is enforced by inferring all possible relations before passing the graph to the next sieve. This best-first architecture is conceptually \final{similar to} L+I but \final{the inference is greedy}, similar to \citet{mani2007three,verhagen2008temporal}.

\final{Although L+I methods impose global constraints in the inference phase, this paper argues that global considerations are necessary in the learning phase as well (i.e., structured learning).
In parallel to the work presented here, \citet{EACL2017StructuredTemporal} also proposed a structured learning approach to extracting the temporal relations. Their work focuses on a domain-specific dataset from Clinical TempEval \citep{bethard2016semeval}, so their work does not need to address some of the difficulties of the general problem that our work addresses. More importantly, they compared structured learning to local baselines, while we find that the comparison between structured learning and L+I is more interesting and important for understanding the effect of global considerations in the learning phase.
In difference from existing methods, we also discuss how to effectively use unlabeled data and how to handle the overwhelming fraction of missing relations in a principled way.  
Our solution targets on these issues and, as we show, achieves significant improvements on two commonly used evaluation sets.}

The rest of this paper is organized as follows. Section~\ref{sec:background} clarifies the temporal relation types and the evaluation metric of a temporal graph used in this paper, Section~\ref{sec:proposed} explains the structured learning approach in detail, and Section~\ref{vaguelinks} discusses the practical issue of missing relations. 
We provide experiments and discussion in Section~\ref{experiments} and conclusion in Section~\ref{conclusion}.

\section{Background}
\label{sec:background}
\subsection{Temporal Relation Types}
Existing corpora for temporal processing often follows the interval representation of events proposed in \citet{allen1984towards}, and makes use of 13 relation types in total.
In many systems, \textit{vague} or \textit{none} is also included as another relation type when a TLINK is not clear or missing.
However, current systems usually use a reduced set of relation types, mainly due to the following reasons.
\begin{enumerate}
	\item The non-uniform distribution of all the relation types makes it difficult to separate low-frequency ones from the others (see Table~1 in \citet{mani2006machine}). For example, relations such as \textit{immediately\_before} or \textit{immediately\_after} barely exist in a corpus compared to \textit{before} and \textit{after}.
	\item Due to the ambiguity in natural language, determining relations like \textit{before} and \textit{immediately\_before} can be a difficult task itself  \citep{chambers2014dense}.
\end{enumerate}
\ignore{In \citet{bethard2007timelines} and \citet{DoLuRo12}, only 5 relations were considered: \textit{before}, \textit{after}, \textit{overlap}, \textit{equal} and \textit{none}; CAEVO \citep{chambers2014dense} uses 6 relations: \textit{before}, \textit{after}, \textit{includes}, \textit{is\_included}, \textit{equal}, and \textit{vague}.
In this work, we follow the reduced set of temporal relation types used in CAEVO because it not only covers all the dominating relations but also has a fine-grained distinction of the relation \textit{overlap}.}
In this work, we follow the reduced set of temporal relation types used in CAEVO \citep{chambers2014dense}: \textit{before}, \textit{after}, \textit{includes}, \textit{is\_included}, \textit{equal}, and \textit{vague}.

\subsection{Quality of A Temporal Graph}
\ignore{The straightforward evaluation metric for a predicted temporal graph is to count the number of correct TLINKs, and compute the precision and recall using it.}
The most recent evaluation metric in TE3, i.e., the temporal awareness \citep{uzzaman2011temporal}, is adopted in this work.
Specifically, let $G_{sys}$ and $G_{true}$ be two temporal graphs from the system prediction and the ground truth, respectively. The precision and recall of temporal awareness are defined as follows.
$$
P = \frac{|G_{sys}^- \cap G_{true}^+|}{|G_{sys}^-|},~R = \frac{|G_{true}^- \cap G_{sys}^+|}{|G_{true}^-|}
$$
where $G^+$ is the closure of graph $G$, $G^-$ is the reduction of $G$, ``$\cap$'' is the intersection between TLINKs in two graphs, and $\lvert G \rvert$ is the number of TLINKs in $G$.
The temporal awareness metric better captures how ``useful'' a temporal graph is. For example, if system 1 produces \event{ripping} is \rel{before} \event{hurt} and \event{hurt} is \rel{before} \event{monitor}, and system 2 adds \event{ripping} is \rel{before} \event{monitor} on top of system 1. Since system 2 is simply a transitive closure of system 1, they would have the same evaluation scores.
Note that \textit{vague} relations are usually considered as non-existing TLINKs and are not counted during evaluation.
\section{A Structured Training Approach}
\label{sec:proposed}
As shown in Fig.~\ref{fig:graph all}, the learning problem in temporal relation extraction is global in nature. 
Even the top local method in TE3, UTTime \citep{laokulrat2013uttime}, only achieved $F_1$=56.5 when presented with a pair of temporal entities (Task C--relation only \citep{uzzaman2013TE3}).
Since the success of an L+I method strongly relies on the quality of the local classifiers, a poor local classifier is obviously a roadblock for L+I methods.
Following the insights from~\citet{PRYZ05}, we propose to use a structured learning approach (also called ``Inference Based Training'' (IBT)).
 
Unlike the current L+I approach, where local classifiers are trained independently beforehand without knowledge of the predictions on neighboring pairs, we train local classifiers with feedback that accounts for other relations, by performing global inference in each round of the learning process. 
In order to introduce the structured learning algorithm, we first explain its most important component, the global inference step.

\subsection{Inference}
\label{sec:inference}
In a document with $n$ pairs of events, let $\phi_i\in\calx\subseteq\real^d$ be the extracted $d$-dimensional feature and $y_i\in\caly$ be the temporal relation for the $i$-th pair of events, $i=1,2,\dots,n$, where $\caly=\{r_j\}_{j=1}^6$ is the label set for the six temporal relations we use.
Moreover, let $\mbf{x}=\{\phi_1,\dots,\phi_n\}\in\calx^n$ and $\mbf{y}=\{y_1,\dots,y_n\}\in\caly^n$ be more compact representations of all the features and labels in this document.
Given the weight vector $\mbf{w}_r$ of a linear classifier trained for relation $r\in\mathcal{Y}$ (i.e., using the one-vs-all scheme), the global inference step is to solve the following constrained optimization problem:
\begin{equation}
\hat{\mbf{y}} = \argmax{\mbf{y}\in\mathcal{C}(\caly^n)}{f(\mbf{x},\mbf{y})},
\label{eq:inference}
\end{equation}
where $\mathcal{C}(\caly^n)\subseteq \caly^n$ constrains the temporal graph to be symmetrically and transitively consistent, and $f(\mbf{x},\mbf{y})$ is the scoring function:
$$
f(\mbf{x},\mbf{y}) = \sum_{i=1}^n{f_{y_i}(\phi_i)}=\sum_{i=1}^n{\frac{e^{\mbf{w}_{y_i}^T\phi_i}}{\sum_{r\in\caly}{e^{\mbf{w}_{r}^T\phi_i}}}}.
$$
\final{Specifically, $f_{y_i}(\phi_i)$ is the probability of the $i$-th event pair having relation $y_i$. $f(x,y)$ is simply the sum of these probabilities over all the event pairs in a document, which we think of as the confidence of assigning $\mbf{y}=\{y_1,...,y_n\}$ to this document and therefore, it needs to be maximized in Eq.~\eqref{eq:inference}.}

Note that when $\mathcal{C}(\caly^n) = \caly^n$, Eq.~\eqref{eq:inference} can be solved for each $\hat{y}_i$ independently, which is what the so-called local methods do, \final{but the resulting $\hat{\mbf{y}}$ may not satisfy global consistency in this way.}
\final{When $\mathcal{C}(\caly^n) \ne \caly^n$, Eq.~\eqref{eq:inference} cannot be decoupled for each $\hat{y}_i$} and is usually formulated as an ILP problem \citep{RothYi04,ChambersJu08,DoLuRo12}.
Specifically, let $\mathcal{I}_r(ij)\in\{0,1\}$ be the indicator function of relation $r$ for event $i$ and event $j$ and $f_r(ij)\in[0,1]$ be the corresponding soft-max score.
Then the ILP objective for global inference is formulated as follows.
\ignore{
\begin{argmaxi}
    {\mathcal{I}}{\sum_{ij\in\mathcal{E}}\sum_{r\in\caly} f_r(ij) \mathcal{I}_r(ij)}
    {}{}
    \addConstraint{\sum_{r}{\mathcal{I}_r(ij)} = 1}{}
    \addConstraint{\mathcal{I}_r(ij) = \mathcal{I}_{\bar{r}}(ji)}{}
    \addConstraint{\mathcal{I}_{r_1}(ij)+\mathcal{I}_{r_2}(jk)-\sum_{m=1}^N \mathcal{I}_{r_{3}^m}(ik)\le 1}{}
\end{argmaxi}}

\begin{eqnarray}
\label{eq:basic ilp}
&\hat{\mathcal{I}} = \textrm{arg}\underset{\mathcal{I}}{\textrm{max}}\sum_{ij\in\mathcal{E}}\sum_{r\in\caly} f_r(ij) \mathcal{I}_r(ij)\\
&\textrm{s.t.}\quad\underset{\textrm{(uniqueness)}}{\Sigma_{r}{\mathcal{I}_r(ij)} = 1},~\underset{\textrm{(symmetry)}}{\mathcal{I}_r(ij) = \mathcal{I}_{\bar{r}}(ji),}\nonumber\\
&\underset{\textrm{(transitivity)}}{\mathcal{I}_{r_1}(ij)+\mathcal{I}_{r_2}(jk)-\Sigma_{m=1}^N \mathcal{I}_{r_{3}^m}(ik)\le 1,}\nonumber
\end{eqnarray}
for all distinct events $i$, $j$, and $k$, where $\mathcal{E}=\{ij~|~\textrm{sentence dist$(i,j)$$\le 1$}\}$, $\bar{r}$ is the reverse of $r$, and $N$ is the number of possible relations for $r_{3}$ when $r_1$ and $r_2$ are true.

\ignore{
\begin{equation}
\label{eq:basic ilp}
\begin{aligned}
&\hat{\mathcal{I}} = \textrm{arg}\underset{\mathcal{I}}{\textrm{max}}
&\sum_{ij\in\mathcal{E}}\sum_{r\in\caly} f_r(ij) \mathcal{I}_r(ij)\\
& \text{s.t.}
&\sum_{r}{\mathcal{I}_r(ij)} = 1\\
&\mathcal{I}_r(ij) = \mathcal{I}_{\bar{r}}(ji)\\
&\mathcal{I}_{r_1}(ij)+\mathcal{I}_{r_2}(jk)-\sum_{m=1}^N \mathcal{I}_{r_{3}^m}(ik)\le 1
\end{aligned}
\end{equation}
subject to $\sum_{r}{\mathcal{I}_r(ij)} = 1$ (uniqueness constraints), $\mathcal{I}_r(ij) = \mathcal{I}_{\bar{r}}(ji)$ (symmetry constraints with $r$ and $\bar{r}$ being reverse relations) and  $\mathcal{I}_{r_1}(ij)+\mathcal{I}_{r_2}(jk)-\sum_{m=1}^N \mathcal{I}_{r_{3}^m}(ik)\le 1$ (transitivity constraints), where $N$ is the number of possible relations for $r_{3}$, for all events $i$ and $j$.}

\begin{figure}[h]
	\centering
	\includegraphics[width=0.45\textwidth]{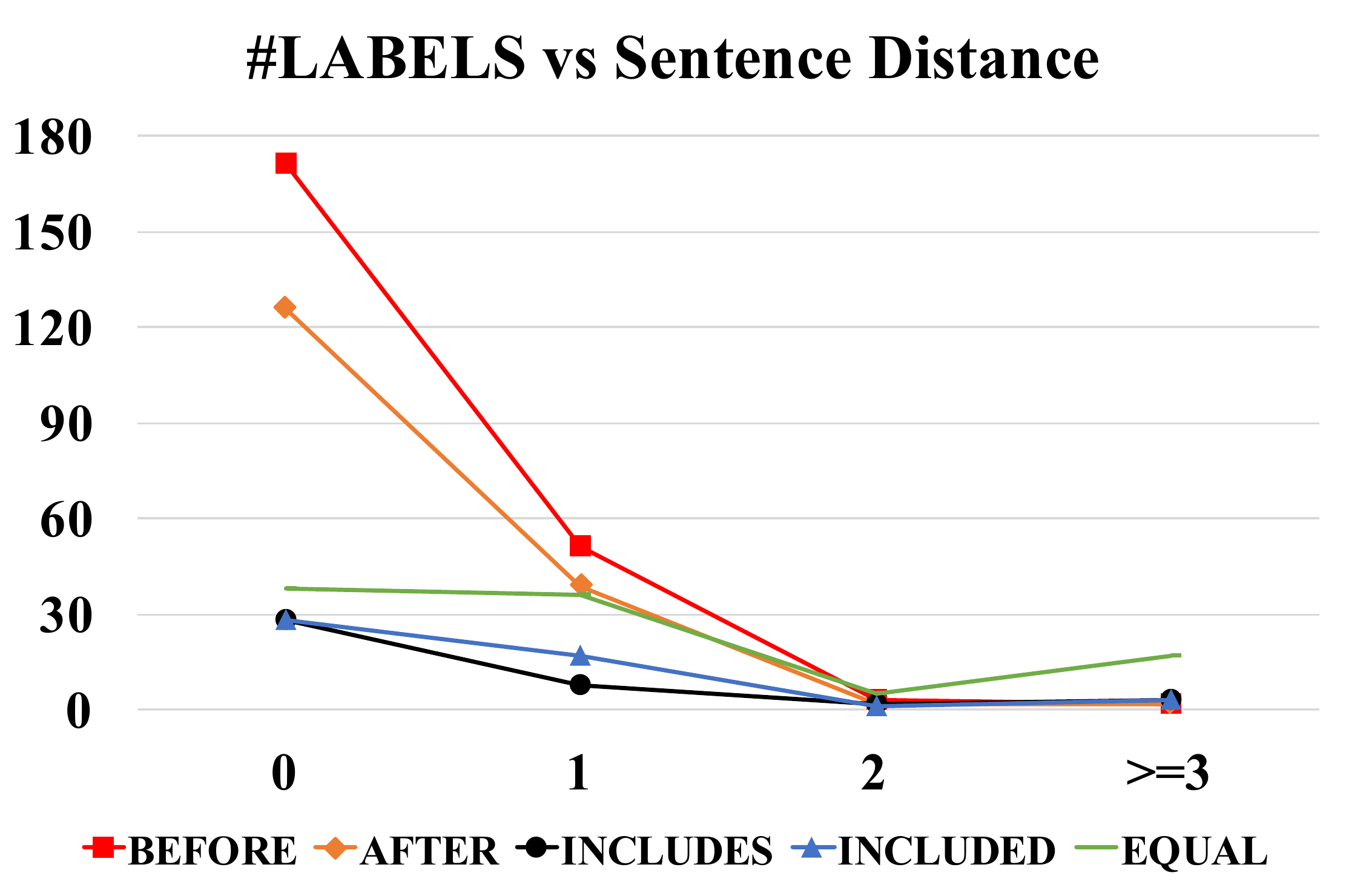}
	\caption{\small \#TLINKs vs sentence distance on the TE3 Platinum dataset. The tail of {\em equal} is due to event coreference and beyond our focus.}
	\label{fig:label_dist}
\end{figure}

\begin{figure}[h]
	\centering
	\includegraphics[width=0.4\textwidth]{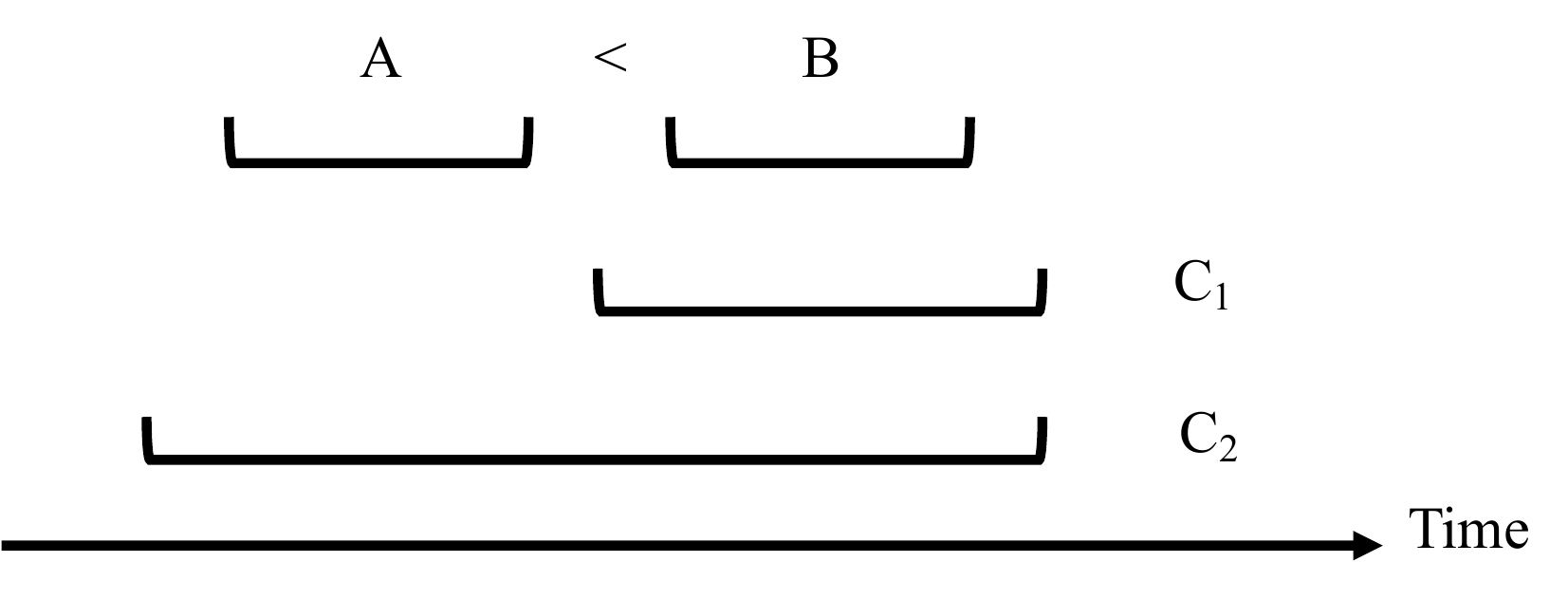}
	\caption{\small When \textit{A is before B} and \textit{B is\_included in C}, \textit{A} can either be \textit{before} \textit{C$_1$} or \textit{is\_included in C$_2$}. We propose to incorporate this via the transitivity constraints for Eq.~\eqref{eq:basic ilp}.}
	\label{fig:transitivity}
\end{figure}

Our formulation in Eq.~\eqref{eq:basic ilp} is different from previous work \citep{ChambersJu08,DoLuRo12} in two aspects: 1) We restrict our event pairs $ij$ to a smaller set \final{$\mathcal{E}=\{ij~|~\textrm{sentence dist$(i,j)$$\le 1$}\}$} where \final{pairs that are} more than one sentence away are deleted for computational efficiency and (usually) for better performance. 
\final{In fact, to make better use of global constraints, we should have allowed more event pairs in Eq.~\eqref{eq:basic ilp}. However, $f_r(ij)$ is usually more reliable when $i$ and $j$ are closer in text. Many participating systems in TE3 \citep{uzzaman2013TE3} have used this {\em pre-filtering} strategy to balance the trade-off between confidence in $f_r(ij)$ and global constraints. We observe that the strategy fits very well to the existing datasets:}
As shown in Fig.~\ref{fig:label_dist}, \final{annotated} TLINKs barely exist if two events are two sentences away.
2) Previously, transitivity constraints were formulated as $\mathcal{I}_{r_1}(ij)+\mathcal{I}_{r_2}(jk)-\mathcal{I}_{r_{3}}(ik)\le 1$, which is a special case when $N=1$ and can be understood as ``$r_1$ and $r_2$ determine a single $r_3$''. However, \final{it was overlooked that, although some $r_1$ and $r_2$ cannot uniquely determine $r_3$, they can still constrain the set of labels $r_3$ can take}.
For example, as shown in Fig.~\ref{fig:transitivity}, when $r_1$=\textit{before} and $r_2$=\textit{is\_included}, $r_3$ is not determined but we know \final{that} $r_3\in\{$\textit{before}, \textit{is\_included}$\}$\footnote{The transitivity table in \citet{allen1983maintaining} shows two more possible relations, {\em overlap} and {\em immediately\_before}, which are not in our label set.}. This information can be easily exploited by allowing $N>1$.

With these two differences, the optimization problem \eqref{eq:basic ilp} can still be efficiently solved using off-the-shelf ILP packages such as GUROBI \citep{Gurobi}.

\subsection{Learning}
With the inference solver defined above, we propose to use the structured perceptron \citep{collins2002} as a representative for the inference based training (IBT) algorithm to learn those weight vectors $\mbf w_r$.
Specifically, let $\mathcal{L}=\{\mathbf{x}_k,\mathbf{y}_k\}_{k=1}^K$ be the labeled training set of $K$ instances (usually documents). The structured perceptron training algorithm for this problem is shown in Algorithm~\ref{algo:sp}. The Illinois-SL package \citep{CSGR10} was used in our experiments for its structured perceptron component.
\final{In terms of the features used in this work, we adopt the same set of features designed for E-E TLINKs in Sec.~3.1 of \citet{DoLuRo12}.}
 
In Algorithm~\ref{algo:sp}, Line~\ref{ln:inference} is the inference step as in Eq.~\eqref{eq:inference} or \eqref{eq:basic ilp}, which is augmented with a closure operation on $\hat{\mathbf{y}}$ in the following line.
In the case in which there is only one pair of events in each instance (thus no structure to take advantage of),  Algorithm~\ref{algo:sp} reduces to the conventional perceptron algorithm and Line~\ref{ln:inference} simply chooses the top scoring label.
With a structured instance instead, Line~\ref{ln:inference} becomes slower to solve, but it can provide valuable information so that the perceptron learner is able to look further at other labels rather than an isolated pair. 
For example in Ex1 and Fig.~\ref{fig:graph all}, the fact that ({ripping},{ordered})=\rel{before}  is established through two other relations: 1) \event{ripping} is an adverbial participle and thus \rel{included} in \event{cascaded} and 2) \event{cascaded} is \rel{before} \event{ordered}. If ({ripping},{ordered})=\rel{before} is presented to a local learning algorithm without knowing its predictions on ({ripping},{cascaded}) and ({cascaded},{ordered}), then the model either cannot support it or overfits it. 
In IBT, however, if the classifier was correct in deciding ({ripping},{cascaded}) and ({cascaded},{ordered}), then ({ripping},{ordered}) would be correct automatically and would not contribute to updating the classifier. 

\begin{algorithm}[htbp!]
	\DontPrintSemicolon 
	\KwIn{Training set $\mathcal{L}=\{\mathbf{x}_k,\mathbf{y}_k\}_{k=1}^K$, learning rate $\lambda$}
	Perform graph closure on each $\mbf{y}_k$\;
	Initialize $\mathbf{w}_r=\mathbf{0}$, $\forall r\in\caly$\;
	\While{convergence criteria not satisfied}{
		Shuffle the examples in $\mathcal{L}$\;
		\ForEach{$(\mathbf{x},\mathbf{y})\in \mathcal{L}$}{
			$\hat{\mathbf{y}} = \textrm{arg}\max_{\mathbf{y}\in\mathcal{C}}{f(\mbf{x},\mbf{y})}$\label{ln:inference}\;
			Perform graph closure on $\hat{\mbf{y}}$\;
			\If{$\hat{\mathbf{y}}\ne \mathbf{y}$}{
				$\mathbf{w}_r =\mbf{w}_r+\lambda (\sum_{i:\mbf{y}_i=r}{\phi_i}-$ $\sum_{i:\hat{\mbf{y}}_i=r}{\phi_i})$, $\forall r\in\caly$\label{ln:weight update}\;
			}
		}
	}
	\Return{$\{\mathbf{w}_r\}_{r\in\caly}$}\;
	\caption{Structured perceptron algorithm for temporal relations}
	\label{algo:sp}
\end{algorithm}
	
\subsection{Semi-supervised Structured Learning}
The scarcity of training data and the difficulty in \final{annotation} have long been \final{a bottleneck} for temporal processing \final{systems}.
Given the inherent global constraints in temporal graphs, we 
propose to perform semi-supervised structured learning using the constraint-driven learning (CoDL) algorithm \citep{ChangRaRo07,ChangRaRo12}, as shown in Algorithm~\ref{algo:codl}, where the function ``Learn'' \final{in} Lines~\ref{ln:learn1} and \ref{ln:learn2} represents any standard learning algorithm (e.g., perceptron, SVM, or even structured perceptron; here we used the averaged perceptron \cite{FreundSc98}) and subscript ``$r$'' means selecting the learned weight vector for relation $r\in\caly$.
CoDL improves the model learned from a small amount of labeled data by repeatedly generating feedback through labeling unlabeled examples, which is in fact a semi-supervised version of IBT.
Experiments show that this scheme is indeed helpful in this problem.

\begin{algorithm}
	\DontPrintSemicolon 
	\KwIn{Labeled set $\mathcal{L}$, unlabeled set $\mathcal{U}$, weighting coefficient $\gamma$}
    Perform closure on each graph in $\mathcal{L}$\;
	Initialize $\mathbf{w}_r=\textrm{Learn}(\mathcal{L})_r, \forall~r\in\caly$\label{ln:learn1}\;
	\While{convergence criteria not satisfied}{
		$\mathcal{T}=\emptyset$\;
		\ForEach{$\mathbf{x}\in \mathcal{U}$}{
			$\hat{\mathbf{y}} = \textrm{arg}\max_{\mathbf{y}\in\mathcal{C}}{f(\mbf{x},\mbf{y})}$\;
			Perform graph closure on $\hat{\mbf{y}}$\;
			$\mathcal{T}=\mathcal{T}\cup \{(\mathbf{x},\hat{\mathbf{y}})\}$\;
		}
		$\mathbf{w}_r =\gamma\mbf{w}_r+(1-\gamma)\textrm{Learn}(\mathcal{T})_r,\forall~r\in\caly$\label{ln:learn2}\label{ln:weight update codl}\;
	}
	\Return{$\{\mathbf{w}_r\}_{r\in\caly}$}\;
	\caption{Constraint-driven learning algorithm}
	\label{algo:codl}
\end{algorithm}

	
\section{Missing Annotations}
\label{vaguelinks}
Since even human annotators find it difficult to annotate temporal graphs, many of the TLINKs are left unspecified by annotators (compare Fig.~\ref{fig:graph} to Fig.~\ref{fig:graph all}).
While some of these missing TLINKs can be inferred from existing ones, the vast majority still \final{remain} unknown as shown in Table~\ref{tab: unspecified}.
Despite the existence of denser annotation schemes (e.g., \citet{cassidy2014annotation}), the TLINK annotation task is quadratic in the number of nodes, and it is practically infeasible to annotate complete graphs. 
Therefore, the problem of identifying these unknown relations in training and test is a major issue that dramatically hurts existing methods.

\begin{table}[htbp!]
	\centering
	\caption{\small Categories of E-E TLINKs in the TE3 Platinum dataset. Among all pairs of events, 98.2\% of them are left unspecified by the annotators. Graph closure can automatically \final{add} 8.7\%, but most of the event pairs are still {unknown}.}
	\label{tab: unspecified}
	\begin{tabular}{ c|c|c|c } 
		\hline
		\multicolumn{2}{c|}{Type} & \#TLINK & \%\\
		\hline
		\multicolumn{2}{c|}{Annotated} & 582 & 1.8\\
		\hline
		\multirow{2}{*}{Missing}& Inferred & 2840 & 8.7\\ & Unknown & 29240 & 89.5\\ 
		\hline
		\multicolumn{2}{c|}{Total} & 32662 & 100\\
		\hline
	\end{tabular}
\end{table}

We could simply use these unknown pairs (or some filtered version of them) to design rules or train classifiers to identify whether a TLINK is {\em vague} or not.
However, we propose to exclude both the unknown pairs and the {\em vague} classifier from the training process -- by changing the structured loss function to ignore the inference feedback on {\em vague} TLINKs (see Line~\ref{ln:weight update} in Algorithm~\ref{algo:sp} and Line~\ref{ln:weight update codl} in Algorithm~\ref{algo:codl}). The reasons are \final{discussed below}.

First, it is believed that a lot of the unknown pairs are not really {\em vague} but rather pairs that the annotators failed to look at \citep{bethard2007timelines,cassidy2014annotation,chambers2014dense}. For example, (cascaded, monitor) should be {\final{annotated as} \rel{before} but is missing in Fig.~\ref{fig:graph}. \final{It is hard to exclude this noise in the data during training}. 
Second, compared to the overwhelmingly large number of unknown TLINKs (89.5\% as \final{shown} in Table~\ref{tab: unspecified}), the scarcity of non-vague TLINKs makes it hard to learn a good {\em vague} classifier.
Third, {\em vague} is fundamentally different from the other relation types. For example, if a {\em before} TLINK can be established given a sentence, then it always holds as {\em before} regardless of other events around it, but if a TLINK is {\em vague} given a sentence, it may still change to other types afterwards if a connection can later be established through other nodes from the context. This distinction emphasizes  that {\em vague} is a consequence of lack of background/contextual information, rather than a concrete relation type to be trained on.
Fourth, without the {\em vague} classifier, the predicted temporal graph tends to become more densely connected, thus the global transitivity constraints can be more effective in correcting local mistakes \citep{ChambersJu08}.
\ignore{
\begin{enumerate}
	\item It is believed that a lot of the {\em vague} TLINKs are actually not vague but with missing annotations \citep{bethard2007timelines,cassidy2014annotation,chambers2014dense}. This noise in data is hard to be excluded from training.
	\item Compared to the overwhelming large number of {\em vague} TLINKs, the scarcity of non-vague TLINKs makes it fundamentally  hard to learn a good {\em vague} classifier.
	\item {Vagueness} is locally undecidable and thus not well-defined for training. For example, if one {\em before} TLINK can be established given a sentence, then it always holds {\em before} regardless of the other sentences, but if a TLINK is {\em vague} according to some sentence, it may still change to other types afterwards if a connection can later be established through nodes from other sentences.
	\item Without the {\em vague} classifier, the predicted temporal graph tends to become more densely connected, for which the global transitivity constraints can be more effective in correcting local mistakes \citep{ChambersJu08}.
\end{enumerate} 
}
    
However, excluding the local classifier for {\em vague} TLINKs would undesirably assign non-vague TLINKs to every pair of events. To handle this, we take a closer look at the {\em vague} TLINKs.
\final{We note that a {\em vague} TLINK could arise in two situations if the annotators did not fail to look at it.}
One is that an annotator looks at this pair of events and decides that multiple relations can exist, and the other one is that two annotators disagree on the relation (similar arguments were also made in \citet{cassidy2014annotation}).
In both situations, the annotators first try to assign all possible relations to a TLINK, and then change the relation to {\em vague} if more than one \final{can be} assigned.
This human annotation process for {\em vague} is different from many existing methods, which either identify the existence of a TLINK first (using rules or machine-learned classifiers) and then classify, or directly include {\em vague} as a classification label along with other non-vague relations.

In this work, however, we propose to mimic this \final{mental} process by a {\em post-filtering} method\footnote{Some systems (e.g., TARSQI \citep{verhagen2008temporal}) \final{employed a} similar idea from a different standpoint, by thresholding TLINKs based on confidence scores.}.
Specifically, \final{we} take each TLINK produced by ILP and determine whether it is {\em vague} using its relative entropy (the Kullback-Leibler divergence) to the uniform distribution. Let $\{r_m\}_{m=1}^{M}$ be the set of relations that the $i$-th pair of events can take, we filter the $i$-th TLINK given by ILP by:     
$$
\delta_i=\sum_{m=1}^M{f_{r_m}(\phi_i) \log{\left(M f_{r_m}(\phi_i)\right)}},
$$
where $f_{r_m}(\phi_i)$ is the soft-max score of $r_m$, obtained by the local classifier for $r_m$.
We then compare $\delta_i$ to a fixed threshold $\tau$ to determine the vagueness of this TLINK; \final{we} accept its originally predicted label if $\delta_i>\tau$, or  change it to {\em vague} otherwise.
Using relative entropy here is intuitively appealing and empirically useful as shown in the experiments section; better metrics are of course yet to be designed.

\ignore{
This process is intuitively appealing and is different to many existing methods, which either identify the existence of a TLINK first (using rules or machine-learned classifiers) and then classify, or directly include {\em vague} as a classification label along with other non-vague relations.
This post-filtering strategy thus provides us with a new perspective to treating {\em vague} TLINKs in temporal relation extraction.
This new perspective towards treating {\em vague} TLINKs gives rise to the {\em post-filtering} method, in contrast to the aforementioned ``pre-filtering'' method.
}

\section{Experiments}
\label{experiments}

\subsection{Datasets}
The TempEval3 (TE3) workshop \citep{uzzaman2013TE3} provided the TimeBank (TB) \citep{pustejovsky2003timebank}, AQUAINT (AQ) \citep{graff2002aquaint}, Silver (TE3-SV), and Platinum (TE3-PT) datasets, where TB and AQ are usually for training, and TE3-PT is usually for testing. 
The TE3-SV dataset is a much larger, machine-annotated and automatically-merged dataset based on multiple systems, with the intention to see if these ``silver'' standard data can help when included in training (although almost all participating systems saw performance drop with TE3-SV included in training).

Two popular augmentations on TB are the Verb-Clause temporal relation dataset (VC) and TimebankDense dataset (TD). 
The VC dataset has specially annotated event pairs that follow the so-called Verb-Clause structure \citep{bethard2007timelines}, which is usually beneficial to be included in training \citep{uzzaman2013TE3}.
The TD dataset contains 36 documents from TB which were re-annotated using the dense event ordering framework proposed in \citet{cassidy2014annotation}.
The experiments included in this paper will involve the TE3 datasets as well as these augmentations. Therefore, some statistics on them are shown in Table~\ref{tab:datasets} for the readers' information.

\ignore{*********************************
\begin{table}[htbp!]
	\centering
	\caption{\small Facts about the datasets used in this paper. The TD dataset is split into train, dev, and test in the same way as in \citet{chambers2014dense}. Note that the column of TLINKs only counts the non-vague TLINKs, from which we can see that the TD dataset has a much higher ratio of \#TLINKs to \#Events.}
	\label{tab:datasets}
	\begin{tabular}{ l|c|c|c|c|c } 
		\hline
		Dataset &Docs 	& Tokens	& Events 	& TLINKs	&	Purpose\\
		\hline
		TB+AQ 	& 256 	& 100K 		& 12K 		& 12K		&	Training\\
		TE3-SV		& 2.5K 	& 666K		&			& \QN{120K}	&	Training\\
		VC		& 132	&			&			& 0.9K		&	Training\\
		TD-Train& 22	&			& 1K		& 7K		&	Training\\
		TD-Dev	& 5		&			& 0.2K		& 1K		&	Development\\
		TD-Test	& 9		&			& 0.4K		& 2K		&	Evaluation\\
		TE3-PT 		& 20 	& 6K 		& 0.7K 		& 0.9K		&	Evaluation\\
		\hline
	\end{tabular}
\end{table}
*********************************}
\begin{table}[htbp!]
	\centering
	\caption{\small Facts about the datasets used in this paper. The TD dataset is split into train, dev, and test in the same way as in \citet{chambers2014dense}. Note that the column of TLINKs only counts the non-vague TLINKs, from which we can see that the TD dataset has a much higher ratio of \#TLINKs to \#Events. The TLINK annotations in TE3-SV is not used in this paper and its number is thus not shown.}
	\label{tab:datasets}
	\begin{tabular}{ l|c|c|c|c } 
		\hline
		Dataset &Doc	& Event 	& TLINK		&	Note\\
		\hline
		TB+AQ 	& 256 	& 12K 		& 12K		&	Training\\
		VC		& 132 	& 1.6K	& 0.9K			&	Training\\
		TD		& 36 	& 1.6K		& 5.7K		&	Training\\
        TD-Train& 22 	& 1K		& 3.8K		&	Training\\
		TD-Dev	& 5	 	& 0.2K		& 0.6K		&	Dev\\
		TD-Test	& 9 	& 0.4K		& 1.3K		&	Eval\\
		TE3-PT 		& 20  	& 0.7K 		& 0.9K		&	Eval\\
		TE3-SV		& 2.5K  & 81K		& -			&	Unlabeled\\
		\hline
	\end{tabular}
\end{table}
	
\subsection{Baseline Methods}
In addition to the state-of-the-art systems, another two baseline methods were also implemented for a better understanding of the proposed ones.
The first is the regularized averaged perceptron (AP) \citep{FreundSc98} implemented in the LBJava package \citep{RizzoloRo10} and \final{is} a local method. 
On top of the first baseline, we performed global inference in Eq.\eqref{eq:basic ilp}, referred to as the L+I baseline (AP+ILP).
Both of them used the same feature set \final{(i.e., as designed in \citet{DoLuRo12})}  as in the proposed structured perceptron (SP) and CoDL for fair comparisons.
To clarify, SP and CoDL are training algorithms and their immediate outputs are the weight vectors {$\{\mathbf{w}_r\}_{r\in\caly}$} for local classifiers. An ILP inference was performed on top of them to yield the final output, and we refer to it as {``S+I''} (i.e., structured learning+inference) methods.
	
\begin{table}[htbp!]
	\centering
	\caption{\small Temporal awareness scores on TE3-PT given gold event pairs. Systems that are significantly better (per McNemar's test with $p<0.0005$) than the previous rows are underlined. The last column shows the relative improvement in F1 score over AP-1, which identifies the source of improvement: 5.2\% from additional training data, 9.3\% (14.5\%-5.2\%) from constraints, and 10.4\% from structured learning.}
	\label{tab: know nones}
	\begin{tabular}{ l|c|c|c|c } 
		\hline
		Method & P & R & F1 & \%\\ 
		\hline
		UTTime & 55.6 & 57.4 & 56.5 & +5.0 \\ 
		\hline
		AP-1 & 56.3 & 51.5 & 53.8 & 0 \\
		\underline{AP-2} & 58.0 & 55.3 & 56.6 & +5.2\\ 
		\underline{AP+ILP} & 62.2 & 61.1 & 61.6 & +14.5 \\
		\hline
		\underline{SP+ILP} & \textbf{69.1} & \textbf{65.5} & \textbf{67.2} & \textbf{+24.9}\\
		\hline
	\end{tabular}
\end{table}

\subsection{Results and Discussion}
\subsubsection{\final{TE3 Task C - Relation Only}}
\ignore{1. given gold non-vague pairs, show the improvement of SP. Need: local results trained on TBAQ}
\ignore{2. without gold non-vague pairs, show the improvement of SP, post-filtering, and CoDL. Need: local results trained on TBAQ; codl+postfilter results}
\ignore{3. compare with caveo. trained on TD-train, compare results on TD-dev and TD-test. Need: results on TD-dev.}
\ignore{4. compare with caveo. trained on TD-train, compare results on TE3-PT}

To show the benefit of using structured learning, we first tested one scenario where the gold pairs of events that have a non-vague TLINK were known priori. This setup was a standard task presented in TE3, so that the difficulty of detecting {\em vague} TLINKs was ruled out. 
\final{This setup also helps circumvent the issue that TE3 penalizes systems which assign extra labels that do not exist in the annotated graph, while these extra labels may be actually correct because the annotation itself might be incomplete.}
UTTime \citep{laokulrat2013uttime} was the top system in this task in TE3. 
Since UTTime is not available to us, and its performance was reported in TE3 in terms of both E-E and E-T TLINKs together, we locally trained an E-T classifier based on \citet{DoLuRo12} and included its prediction only for fair comparison.

\begin{table*}[htbp!]
	\centering
	\caption{\small Temporal awareness scores given gold events but with no gold pairs, which show that the proposed S+I methods outperformed state-of-the-art systems in various settings. The fourth column indicates the annotation sources used, with additional unlabeled dataset in the parentheses. The ``Filters'' column shows if the pre-filtering method (Sec.~\ref{sec:inference}) or the proposed post-filtering method (Sec.~\ref{vaguelinks}) were used. The last column is the relative improvement in $F_1$ score compared to baseline systems on line 1, 7, and 11, respectively. Systems that are significantly better than the ``*''-ed systems are underlined (per McNemar's test with $p<0.0005$).}
	\label{tab: dont know nones}
	\small
	\begin{tabular}{ c|c|c|c|c|c|c|c|c|c } 
		\hline
		No. & System & Method & Anno. (Unlabeled) & Testset & Filters	& P 	& R 	& F1 	& \%\\ 
		\hline
        \hline
		1& ClearTK & Local & TB, AQ, VC, TD &TE3-PT& pre & \textbf{37.2} 	& 33.1 	& 35.1 	& 0  \\ 
		2& AP* & Local & TB, AQ, VC, TD&TE3-PT& pre	& 35.3	& 37.1  & 36.1	& +2.8\\
		3& AP+ILP & L+I&	TB, AQ, VC, TD&TE3-PT& pre & 35.7 	& 35.0 	& 35.3 	& +0.6\\
		4& \underline{SP+ILP} 	& S+I &TB, AQ, VC, TD&TE3-PT& pre	& 32.4 	& 45.2 	& 37.7 	& +7.4 \\
		5& \underline{SP+ILP} &S+I & TB, AQ, VC, TD&TE3-PT& pre+post
				& 33.1 	& \textbf{49.2} 	& 39.6 	& +12.8\\
		6& \underline{CoDL+ILP} &S+I & TB, AQ, VC, TD (TE3-SV)&TE3-PT& pre+post
				& 35.5 	& 46.5 	& \textbf{40.3} & \textbf{+14.8}\\
		\hline
        7& ClearTK* & Local & TB, VC &TE3-PT& pre &  \textbf{35.9}	&  38.2	& 37.0	&  0\\  
        8& \underline{SP+ILP} &S+I & TB, VC&TE3-PT& pre+post
				& 30.7& \textbf{47.1} 	& 37.2   & +0.5 \\
        9& \underline{CoDL+ILP} &S+I & TB, VC (TE3-SV)&TE3-PT& pre+post
				&33.9  	&45.9  	&\textbf{39.0}  & \textbf{+5.4}\\
		\hline\hline
		10&ClearTK&Local&TD-Train&TD-Test&pre&46.04& 20.90 & 28.74&-\\
		11&CAEVO*&L+I&TD-Train&TD-Test&pre&\textbf{54.17}& 39.49 & 45.68 & 0\\
		12&\underline{SP+ILP}&S+I&TD-Train&TD-Test&pre+post&45.34 & 48.68 &46.95&+3.0\\
		13&\underline{CoDL+ILP}&S+I&TD-Train (TE3-SV)&TD-Test&pre+post& 45.57 & \textbf{51.89} & \textbf{48.53}& \textbf{+6.3}\\
		\hline
	\end{tabular}
\end{table*}

UTTime is a local method and was trained on TB+AQ and tested on TE3-PT. We used the same datasets for our local baseline and its performance is shown in Table~\ref{tab: know nones} under the name ``AP-1''. Note that the reported numbers below are the temporal awareness scores obtained from the official evaluation script provided in TE3. We can see that UTTime is about 3\% better than AP-1 in the absolute value of $F_1$, which is expected since UTTime included more advanced features derived from syntactic parse trees.
By adding the VC and TD datasets into the training set, we retrained our local baseline and achieved comparable performance to UTTime (``AP-2'' in Table~\ref{tab: know nones}). On top of AP-2, a global inference step enforcing symmetry and transitivity constraints (``AP+ILP'') can further improve the $F_1$ score by 9.3\%, which is consistent with previous observations \citep{ChambersJu08, DoLuRo12}. SP+ILP further improved the performance in precision, recall, and $F_1$ significantly (per the McNemar's test \citep{everitt1992analysis,dietterich1998approximate} with $p<$0.0005), reaching an $F_1$ score of 67.2\%. This meets our expectation that structured learning can be better when the local problem is difficult \citep{PRYZ05}. 

\subsubsection{\final{TE3 Task C}}
In the first scenario, we knew in advance which TLINKs existed or not, so the ``pre-filtering'' (i.e., ignoring distant pairs as mentioned in Sec.~\ref{sec:inference} and ``post-filtering'' methods were not used when generating \final{the results in} Table~\ref{tab: know nones}.
We then tested a more practical scenario, where we only knew the events, but did not know which ones are related.
This setup was Task C in TE3 and the top system was ClearTK \citep{bethard2013cleartk}. Again, for fair comparison, we simply added the E-T TLINKs predicted by ClearTK. Moreover, 10\% of the training data was held out for development. Corresponding results on the TE3-PT testset are shown in Table~\ref{tab: dont know nones}.

From lines 2-4, all systems see significant drops in performance if compared with the same entries in Table~\ref{tab: know nones}. It confirms our assertion that how to handle {\em vague} TLINKs is a major issue for this temporal relation extraction problem.
The improvement of SP+ILP (line 4) over AP (line 2) was small and AP+ILP (line 3) was even worse than AP, which necessitates the use of a better approach towards {\em vague} TLINKs.
By applying the post-filtering method proposed in Sec.~\ref{vaguelinks}, we were able to achieve better performances using SP+ILP (line 5), which shows the effectiveness of this strategy.
Finally, by setting $\mathcal{U}$ in Algorithm~\ref{algo:codl} to be the TE3-SV dataset, CoDL+ILP (line 6) achieved the best $F_1$ score with a relative improvement over ClearTK being 14.8\%.
Note that when using TE3-SV in this paper, we did not use its annotations on TLINKs because of its well-known large noise \citep{uzzaman2013TE3}.

In \citet{uzzaman2013TE3}, we notice that the best performance of ClearTK was achieved when trained on TB+VC (line 7 is higher than its reported values in TE3 because of later changes in ClearTK), so we retrained the proposed systems on the same training set and results are shown on lines 8-9.
In this case, the improvement of S+I over Local was small, which may be due to the lack of training data. Note that line 8 was still significantly different to line 7 per the McNemar's test, although there was only 0.2\% absolute difference in $F_1$, which can be explained from their large differences in precision and recall.

\subsubsection{\final{Comparison with CAEVO}}
The proposed structured learning approach was further compared to a recent system, a CAscading EVent Ordering architecture (CAEVO) proposed in \citet{chambers2014dense} (lines 10-13).
We used the same training set and test set as CAEVO in the S+I systems. Again, we added the E-T TLINKs predicted by CAEVO to both S+I systems.
In \citet{chambers2014dense}, CAEVO was reported on the straightforward evaluation metric including the {\em vague} TLINKs, but the temporal awareness scores were used here, which explains the difference between line 11 \final{in Table~\ref{tab: dont know nones}} and \final{what was reported in} \citet{chambers2014dense}.

ClearTK was reported to be outperformed by CAEVO on TD-Test \citep{chambers2014dense}, but we observe that ClearTK on line 10 was much worse even than itself \final{on line 7 (trained on TB+VC) and on line 1 (trained on TB+AQ+VC+TD)} due to the annotation scheme difference between TD and TB/AQ/VC. 
\final{ClearTK was designed mainly for TE3, aiming for high precision, which is reflected by its high precision on line 10, but it does not have enough flexibility to cope with two very different annotation schemes.}
Therefore, we have chosen CAEVO as the baseline system to evaluate the significance of the proposed ones.
On the TD-Test dataset, all systems other than ClearTK had better $F_1$ scores compared to their performances on TE3-PT. This notable difference (i.e., 48.53 vs 40.3) indicates the better quality of the dense annotation scheme that was used to create TD \citep{cassidy2014annotation}.
SP+ILP outperformed CAEVO and if additional unlabeled dataset TE3-SV was used, CoDL+ILP achieved the best score with a relative improvement in $F_1$ score being 6.3\%.

We notice that the proposed systems often have higher recall than precision, and that this is less an issue on a densely annotated testset (TD-Test), so their low precision \final{on TE3-PT} possibly came from the missing annotations on TE3-PT. It is still under investigation how to control precision and recall in real applications.

\section{Conclusion}
\label{conclusion}

We develop a structured learning approach to identifying temporal relations in natural language text and show that it \final{captures the global nature of this problem better} than state-of-the-art systems do. 
A new perspective towards {\em vague} relations is also proved to \final{gain from} 
fully taking advantage of the structured approach.
In addition, the global nature of this problem gives rise to a better way \final{of} making use of the readily available unlabeled data, which further improves the proposed method.
\final{The improved performance on both TE3-PT and TD-Test, two 
differently annotated datasets, clearly shows the advantage of the proposed method over existing methods.}
We plan to build on the notable improvements shown \final{here}
and expand this study to deal with additional temporal reasoning problems in natural language text. 

\section*{Acknowledgements}
\final{We thank all the reviewers for providing useful comments. This
research is supported in part by a grant from the Allen Institute for Artificial Intelligence (allenai.org); the IBM-ILLINOIS Center for Cognitive Computing Systems Research (C3SR) - a research collaboration as part of the IBM Cognitive Horizon Network; by the US Defense Advanced Research Projects Agency (DARPA) under contract FA8750-13-2-0008; and by the Army Research Laboratory (ARL) under agreement W911NF-09-2-0053.
The views and conclusions contained herein are those of the authors and should not be interpreted as necessarily representing the official policies of the U.S. Government.}

	
	\bibliography{emnlp2017,cited-long,ccg-long}
	\bibliographystyle{emnlp_natbib}
	
\end{document}